\documentclass[runningheads]{waica}

\usepackage{fontspec}
\usepackage{graphicx}
\usepackage{microtype}
\usepackage{url}
\usepackage{booktabs}
\usepackage{amsmath}
\usepackage{amssymb}
\usepackage{soul}
\usepackage{makecell}
\usepackage{multirow}
\usepackage{tabularx}
\usepackage{algorithm}
\usepackage{algorithmic}
\usepackage[table]{xcolor}
\usepackage[most]{tcolorbox}
\usepackage{placeins}
\usepackage[colorinlistoftodos,textsize=tiny]{todonotes}
\usepackage[skip=5pt]{caption}
\usepackage{subcaption}
\usepackage{wrapfig}
\usepackage{siunitx}
\usepackage{listings}
\usepackage[numbers,sort&compress]{natbib}
\usepackage[hidelinks]{hyperref}

\newtcblisting{PromptBox}[1]{%
    breakable,
    enhanced,
    listing only,
    title={#1},
    listing engine=listings,
    colback=black!4,
    colframe=black,
    colbacktitle=black!85,
    coltitle=white,
    fonttitle=\bfseries,
    listing options={
        breaklines=true,
        breakautoindent=false,
        breakindent=0pt,
        basicstyle=\normalfont,
        columns=fullflexible,
        keepspaces=true,
        escapeinside={(*@}{@*)}
    }
}

\newcommand{\method}{\textsc{SimPref}}
\newcommand{\Sref}[1]{\S\ref{#1}}

\begin{document}

\title{Step-Level Preference Learning for Generative Agents in Social Simulations}
\titlerunning{Step-Level Preference Learning for Generative Agents}

% Use dagger (\dagger, \ddagger) instead of star for title-page footnotes
\makeatletter
\renewcommand{\@fnsymbol}[1]{\ensuremath{\ifcase#1\or \dagger\or \ddagger\or \star\else\@ctrerr\fi}}
\makeatother
% Separate authors with commas only (no "and" before the last author)
\renewcommand{\lastandname}{\unskip,}

\author{Wenchang Gao\inst{1}\thanks{Work done during an internship at Shanghai Qi Zhi Institute. Email: \email{wenchanggao93@gmail.com}} \and
Pingyue Sheng\inst{3} \and
Lanlan Qiu\inst{1} \and
Yunfei Ma\inst{1} \and
Jian Zhao\inst{2} \and
Baicheng Chen\inst{2,4} \and
Kangda Wang\inst{2} \and
Yuyang Tian\inst{1,5} \and
Shunqiang Mao\inst{1,6} \and
Tianxing He\inst{3,2,1}\thanks{Corresponding author. Email: \email{hetianxing@mail.tsinghua.edu.cn}}}

\authorrunning{W. Gao et al.}

% Affiliations flow as a filled, wrapping paragraph (not one per line).
% Superscripts are set manually and must match each author's \inst{} number.
\institute{%
$^{1}$Shanghai Qi Zhi Institute \quad
$^{2}$Xiongan AI Institute \quad
$^{3}$Tsinghua University \quad
$^{4}$CUHK-Shenzhen \quad
$^{5}$USTC \quad
$^{6}$Sun Yat-sen University}

\maketitle

\begin{abstract}
    Large language model (LLM)-based generative agents simulate human behavior through long-horizon decision-making processes that comprise intermediate steps such as planning, memory retrieval, reflection, and action selection. However, fine-grained human annotations of these intermediate steps remain scarce, and existing agents are not grounded in human preferences over such intermediate decisions. To address this gap, we introduce \method, an interactive simulation interface that enables us to collect step-level human preference supervision over agent decision trajectories, leading to a dataset of 57K fine-grained annotations. We conduct step-level preference learning on open-weight language models using supervised finetuning and direct preference optimization on this data, consistently improving simulation fidelity, coordination, and interaction quality, and inducing more socially effective agent behavior. Our results show that step-level human supervision is an effective training signal for improving both local decision quality and long-horizon agent behavior. The related code and dataset will be open-sourced in \url{https://github.com/W-GaoT/SimPref}.

{\footnotesize
}
\end{abstract}

\vspace{-10pt}
\section{Introduction}
\label{sec:intro}
\vspace{-10pt}

Simulating realistic human behavior in complex social contexts is a long-standing problem in cognitive science and artificial intelligence \citep{squazzoni2014social,park2024simulations1000}. Starting from Generative Agents (GA) \citep{park2023generative}, systems that combine a language model with a structured agent architecture composed of modules for memory retrieval, reflection, planning, and action, have enabled increasingly complex social worlds in which agents maintain personas, react to events, and exhibit long-horizon routines \citep{yu2024affordable,chen2024towards,feng2025simcity,tian2025visualized}.

% Despite the progress, the intermediate decision modules underlying agent behaviors are poorly understood. fine-grained human annotations of intermediate module outputs are largely absent, and the architectural modules underlying these agents are rarely grounded in step-level data derived from real human preferences. Most existing works either build new architectural variants \citep{yu2024affordable,wang2023humanoid,chen2024towards}, or apply agent simulations in diverse social contexts \citep{li2024econagent,cau2025language,tian2025visualized,li2025metaagents}. 
% In both cases, human annotation is typically collected post hoc using trajectory-level judgments, such as Likert-scale ratings after simulation completion, yielding only sparse supervision of the final behavior rather than of the intermediate decisions that produced it. 
% Such holistic annotations make it difficult to identify which intermediate decisions led to downstream failures and therefore provide little actionable supervision for improving individual decision modules. 

Despite this progress, the fine-grained processes underlying agent behavior remain poorly understood. While language agents can produce increasingly realistic long-horizon behaviors, we still have a limited understanding of how their intermediate decisions are formed and where failures emerge. Existing work has largely focused either on proposing new architectural variants \citep{yu2024affordable,wang2023humanoid,chen2024towards} or on applying agent simulations to diverse social settings \citep{li2024econagent,cau2025language,tian2025visualized,li2025metaagents}. In both lines of work, human annotation is typically collected only after the simulation ends, such as trajectory-level judgments or Likert-scale ratings over complete behaviors. 
Relative to the intermediate decisions that produce behavior, such supervision is coarse-grained and is only available after the simulation ends: it evaluates the final outcome but offers limited signal about how that outcome was produced. This granularity gap makes it difficult to attribute downstream failures to specific local decisions, or to derive fine-grained, actionable supervision for improving them.

To address this gap, this work introduces \method, an interactive interface for step-level human supervision in GA-style social simulation. 
For several human-controlled agents, \method\ exposes their intermediate behaviors to annotators, who can supervise agent behavior by selecting or customizing their preferred outcome. 
Leveraging this interface, we construct the first step-level human preference dataset for GA-style social simulation. 
Our dataset contains 57,239 preference pairs spanning six decision modules: planning, importance scoring, reflection question generation, reflection, action, and dialogue. 
Our dataset is collected from 30 designed social events, each involving 4–8 agents with individual goals under partial observability.

% To enable efficient step-level human supervision of agent decision making, we introduce an interactive social simulation interface, \emph{\method}. 
% The interface reveals the outputs of several human-controlled agents to annotators, allowing them to guide the agents' behavior by selecting or customizing their preferred outcomes. 
% Each agent runs in its own worker thread, while the simulation engine maintains the centralized world state and clock, enabling efficient, user-friendly supervision of simulations. 

Our empirical evaluation shows that step-level preference learning substantially improves GA-style social simulation quality. We perform step-level preference learning on open-weight base language models using supervised finetuning (SFT) and direct preference optimization (DPO) \citep{rafailov2024dpo}. On 10 held-out social events, preference learning consistently enhances open-weight LLMs across whole-trajectory metrics. 
Our analysis also shows that preference-aligned agents allocate their time more properly across activities. 
We further provide case studies that illustrate how these gains manifest in concrete decisions. 
These results demonstrate that step-level human supervision is an effective training signal for improving both local decision quality and long-horizon agent behavior.

Our contributions are summarized as follows:
\begin{itemize}
    \item We introduce \method, an interactive social simulation interface that enables human guidance of step-level outputs from agent modules.
    \item We build the first step-level human preference dataset conditioned on a GA-style architecture, spanning the agent's internal decision process and behavior generation across a diverse range of social scenarios.
    \item We conduct extensive experiments and demonstrate that step-level human supervision is an effective training signal for improving both local decision quality and long-horizon agent behavior.
\end{itemize}

\vspace{-10pt}
\section{Related Work}
\label{sec:related_work}
\vspace{-10pt}

\paragraph{Agent-based social simulation.}
Recent work on LLM agents has popularized modular architectures that decompose behavior into modules such as memory retrieval, reflection, planning, and action, enabling long-horizon simulation of daily routines and social interactions \citep{park2023generative,yu2024affordable,yuan2025evoagent,li2024evolving,li2025metaagents}. Building on this paradigm, many follow-up systems extend the architecture (e.g., with basic needs \citep{wang2023humanoid}, cost reduction \citep{yu2024affordable,mou2025ecolang}, intrinsic desires \citep{wang2025simulating}, and long-term values \citep{chen2024towards}), or port it to new environments, like urban mobility \citep{ye2025mobilecity,bougie2025citysim}, economy simulation \citep{li2024econagent,feng2025simcity,tomasev2025virtual}, or opinion evolution \citep{hou2025society,cau2025language}. 
% These works report improved realism or task success under scenario-specific metrics \citep{gao2025s3}. 
These works primarily evaluate agents at the holistic level, leaving a gap in the grounding and validation of the intermediate module outputs that actually drive behavior. 

\paragraph{Interfaces and benchmarks for social simulation.}
A parallel line of work provides interfaces and environments to standardize agent evaluation and enable scalable experiments~\citep{wang2023humanoid,wang2025yulanonesim,chen2023agentverse,tian2025visualized,piao2025agentsociety}. Such interfaces emphasize scenario design and systematic comparisons across models and settings. 
However, these interfaces do not provide direct human supervision over step-level module outputs during simulation. 
In contrast, \method \ is designed specifically to support direct human intervention in intermediate agent decisions in GA-style social simulation, enabling supervision at the level where downstream behavior is formed.

\paragraph{Human-in-the-loop supervision and preference learning.}
Human feedback has long been used to align policies in the LLM post-training pipeline \citep{liu2023training,shao2024grpo,rafailov2024dpo,movva2025whats}. Existing works in human preference data collection and alignment typically focus on instruction following \citep{ouyang2022rlhf,ji2024alignanythingtrainingallmodality}, securing model harmlessness \citep{bai2022training,ji2025pkusafe,tan2025equilibrate}, or stylistic compliance via preference modeling \citep{dong2023steerlm,wang2024helpsteer2,wang2025helpsteer3}. 
In contrast, our work studies preference learning in interactive social simulation, where feedback is attached to triggered module outputs under partial observability and multi-agent interaction. 
To our knowledge, fine-grained human preference data for GA-style simulation at this level has not been previously studied. 
Our work fills this gap by introducing step-level human supervision and using it to align intermediate agent decisions in social simulation.
% Fine-grained preference studies and datasets for GA-style simulation remain missing in the field. 
% Our study aims to address this gap by operationalizing preference learning for simulation at the step level with real human preference data. 
% Our work differs in both the unit of supervision and the conditioning: we collect goal-conditioned normative preferences at each triggered cognitive component (planning, retrieval, reflection, importance, talk/action), enabling component-wise SFT/DPO and ablations between \emph{direct} (planning/action) and \emph{indirect} (importance/retrieval/reflection) alignment.

% \paragraph{Evaluation of agent behavior.}
% Evaluation in social simulation remains challenging; many works rely on LLM-as-a-judge or human judgments over full trajectories, measuring plausibility, coherence, or social realism~\citep{park2023generative_agents,agentsociety2025,yulanonesim2024}. Such evaluations provide holistic signals but often under-determine the causes of failure and are vulnerable to confounds (e.g., style vs feasibility). Our evaluation combines (i) trajectory-level human-interpretable metrics (spatiotemporal adherence and goal/interaction metrics) with (ii) judge-independent compliance statistics derived from engine preconditions (observability, busy-target exclusivity, and location validity). This dual view allows us to connect component-level preference mismatch to concrete failure modes in grounded behavior.

\vspace{-10pt}
\section{Preliminary}
\label{sec:preliminary}
\vspace{-10pt}

To simulate human behavior, prior work~\citep{park2023generative} proposes a generative agent architecture comprising multiple modules: in the simulation environment, the agent \textit{perceives} recent events, stores them as \textit{memories}, which are high-level descriptions of the perceptions in natural language. To perform any action, the agent \textit{retrieves} a small set of relevant memories, \textit{reflects} based on the retrieved memories, and \textit{plans} for the near future. 

In this work, we study social simulation in multi-agent social event settings.
Each simulation contains agents that must host, attend, coordinate, and interact around an event.
Each agent $i$ is assigned an explicit event-related goal $g_i$ that specifies what it is trying to accomplish.
At time step $t$, agent $i$ receives a partial observation set $o_{i,t}$, consisting of events within its local perceptual range.
The agent converts these observations into memory entries and stores them in its memory bank $\mathcal{M}_i$.
Each memory $m \in \mathcal{M}_i$ is associated with an LLM-estimated importance score.

During memory retrieval, the agent forms a context-dependent query $q_t$ and retrieves a small subset of memories $\mathcal{R}_{i,t} \subset \mathcal{M}_i$ according to a weighted retrieval score:
\begin{equation*}
\mathrm{score}(m; q_t)=\alpha\,\mathrm{rel}(m,q_t)+\beta\,\mathrm{recency}(m)+\gamma\,\mathrm{imp}(m),
\end{equation*}
where $\mathrm{rel}(m,q_t)$ measures semantic relevance between memory $m$ and the current query, $\mathrm{recency}(m)$ favors recently observed information, and $\mathrm{imp}(m)$ denotes the memory's importance score. $\alpha$, $\beta$, and $\gamma$ weight the three terms.

The retrieved memories are then used by several downstream modules. 
The \emph{plan} module produces short-horizon intentions or schedules that help the agent organize its next steps in the event. 
The \emph{reflect} module synthesizes multiple memories into higher-level insights, while the \emph{reflection questions} module generates targeted queries that guide how reflection should occur. 
The \emph{talk} module produces utterances for social interaction, and the \emph{act} module selects actions in the environment, such as moving, approaching another agent, or participating in an activity. 
Together, these modules define the architecture backbone of the agents studied in this work. 
% Detailed descriptions of each module are provided in Appendix~\ref{app:modules}.

% \abe{explain in more details what these cognitive components are (like making readers have a sense of how they are implemented). Also, be consistent about using emph or texttt}
% Agents further \emph{plan} high-level arrangements for future, occasionally \emph{reflect} on important memories retrieved with \emph{reflection questions} query to synthesize memories into long-term insights, and finally \emph{act} in the environment.

\vspace{-10pt}
\section{Framework}
\label{sec:framework}
\vspace{-10pt}

\begin{figure}[t]
    \centering
    \includegraphics[width=\linewidth]{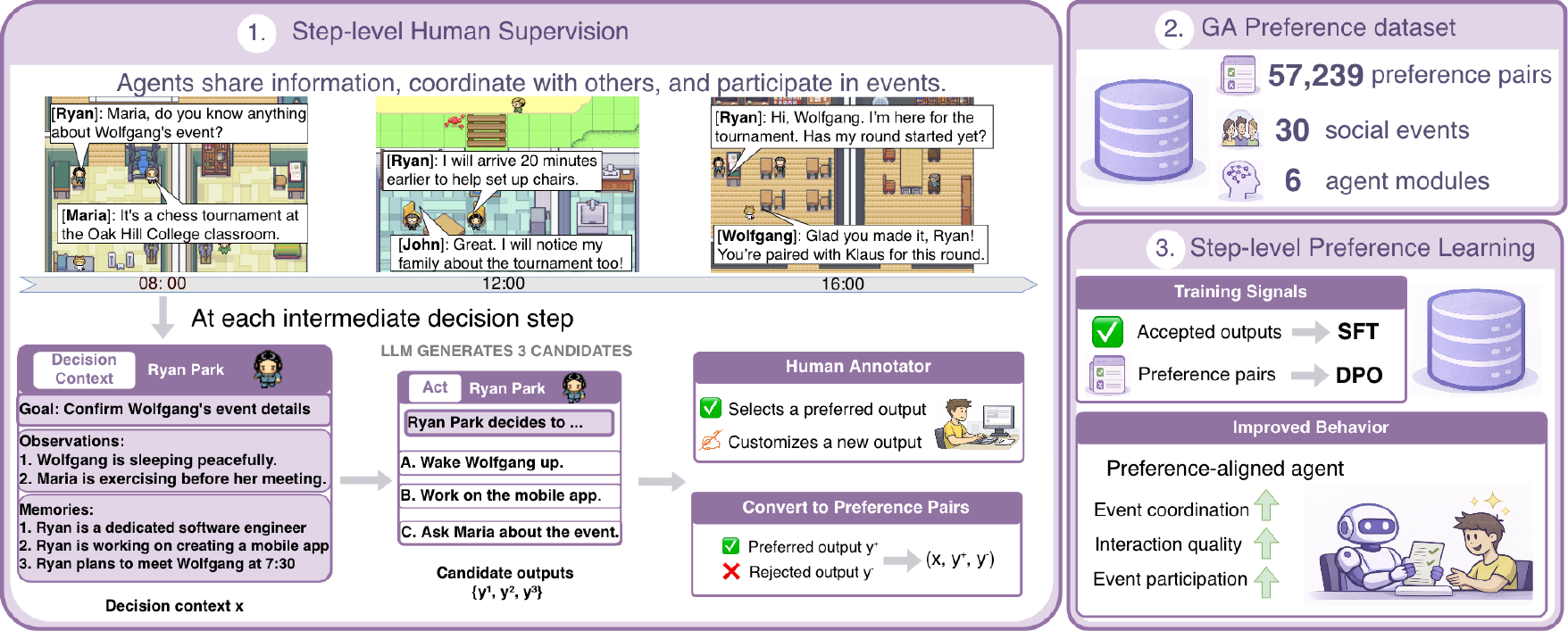}
    % \caption{Overview of our framework. (1) \method \ enables efficient step-level human guidance throughout GA simulation. At each cognitive step, a decision context is constructed, and LLM generates 3 diverse candidates for annotator selection. A human annotator selects a preferred output or customizes a new output, resulting in one accepted output and multiple rejected outputs, stored as pairwise preference tuples in our dataset. (2) Repeating this process produces a GA preference dataset with 57,239 preference pairs spanning 30 social events and 6 cognitive components. (3) Accepted outputs are used for supervised fine-tuning (SFT), while pairwise preferences $(x,y^+,y^-)$ are used for direct preference optimization (DPO). Step-level finetuning produces preference-aligned cognition and improved event coordination, interaction quality, and event participation in simulations.}
    \caption{Overview of our framework. (1) \method \  enables step-level human supervision. At each triggered decision step, the LLM backend of the agent generates three candidate outputs from the current decision context, and a human annotator selects a preferred output or provides a customized alternative. This yields one accepted output and multiple rejected outputs, stored as pairwise preference tuples in the dataset. (2) Repeating this process produces a GA preference dataset with 57,239 preference pairs spanning 30 social events and 6 agent modules. (3) Accepted outputs are used for supervised fine-tuning, while pairwise preferences $(x, y^{+}, y^{-})$ are used for direct preference optimization. }
    \label{fig:framework}
    \vspace{-1.0em}
\end{figure}

We formulate GA-style social simulation at the level of triggered agent internal modules, where each module produces a specific output given the agent’s current local context. 
This formulation turns agent execution into a sequence of step-level decisions that can be directly supervised by human preferences. 
As shown in Figure~\ref{fig:framework}, our framework has three parts: (1) an interactive interface for step-level human annotation, (2) a preference dataset built from these annotations, and (3) a preference learning pipeline that uses the resulting data to align agent behavior with human expectations.
\vspace{-2mm}

\vspace{-6pt}
\subsection{Simulator Interface}
\label{sec:interface}
\vspace{-6pt}
\method \ is an interactive interface for collecting step-level human preferences from GA-style simulations. During the simulation, annotators control 2--3 agents.\footnote{The number of human-controlled agents is set to 2--3 as a balance between event coverage and annotation burden. A single controlled agent often cannot adequately influence or participate in the event, whereas controlling too many agents makes annotation difficult.} Whenever a module of these human-controlled agents is triggered, the interface reveals the same decision context provided to the model, including the agent's partial observations, explicit goal, retrieved memories, and local state. The model then generates three diverse candidate outputs, and the annotator either selects the preferred one or provides a customized alternative. Rather than subjective social taste, annotators judge \emph{decision competence}, i.e., the feasibility of the output, its consistency with the current information state, its temporal and spatial appropriateness, and whether it advances the agent's explicit goal, using only the goal, retrieved memories, and observations exposed in the same decision context given to the model. Interface illustrations and implementation details are provided in Appendices~\ref{app:interface} and~\ref{app:details}.
\vspace{-1mm}
\vspace{-6pt}
\subsection{Preference Collection Pipeline}
\label{sec:data_pipeline}
\vspace{-6pt}

We collect training data from 30 social events, each involving 4--8 agents with distinct goals under partial observability.
The scenarios are designed to distribute key event information asymmetrically across agents, thereby requiring information exchange and coordination through social interaction. We provide detailed event design principles in Appendix~\ref{app:details}.

For each triggered module $k$ of agent $i$ at simulation step $t$, we construct a decision context
$x^{(k)}_{i,t} = \bigl(o_{i,t}, g_i, \mathcal{R}_{i,t}, z_{i,t}\bigr)$,
where $o_{i,t}$ denotes the current partial observations, $g_i$ the explicit goal, $\mathcal{R}_{i,t}$ the retrieved memories, and $z_{i,t}$ the local state features. 
Given $x^{(k)}_{i,t}$, base model GPT-4o \citep{openai_gpt4o_2024} generates three candidate outputs $\{y^{(1)}, y^{(2)}, y^{(3)}\}$. A human annotator selects the most preferred candidate or provides a customized alternative.
We store the final accepted output together with the rejected candidates as step-level preference records, which can be decomposed into pairwise tuples $\bigl(x^{(k)}_{i,t}, y^+, y^-\bigr)$, 
where $y^+$ denotes the human-preferred output and $y^-$ a rejected alternative. 
A customized output is treated as the accepted output and paired with all model-generated candidates as rejected alternatives. 

\vspace{-6pt}
\subsection{Step-level Preference Learning}
\label{sec:pref_setup}
\vspace{-6pt}

We formulate learning step-level human preferences for GA-style simulation as the problem of learning a module-conditioned backbone language model $\pi_\theta(y \mid x, k)$,
where $x$ is the decision context and $k$ denotes the triggered agent module.
We perform step-level preference learning of the base models in two stages.
In the first stage, we apply supervised finetuning (SFT) on the human-accepted outputs: for each record we maximize the likelihood of the accepted output $y^+$ conditioned on its decision context $x$ and triggered module $k$.
In the second stage, we apply Direct Preference Optimization (DPO) \citep{rafailov2024dpo} to the accepted--rejected pairs to further align the policy with human preferences. 
% The SFT and DPO objectives are:
% \[
% \begin{aligned}
% \mathcal{L}_{\mathrm{SFT}}(\theta)
% &=
% -
% \sum_{(x,k,y^+) \in \mathcal{D}_{\mathrm{SFT}}}
% \log \pi_\theta(y^+ \mid x, k), \\
% \mathcal{L}_{\mathrm{DPO}}(\theta)
% &=
% -
% \sum_{(x,k,y^+,y^-) \in \mathcal{D}_{\mathrm{DPO}}}
% \log \sigma \!\left(
% \beta \log \frac{\pi_\theta(y^+ \mid x,k)}{\pi_\theta(y^- \mid x,k)}
% -
% \beta \log \frac{\pi_{\mathrm{ref}}(y^+ \mid x,k)}{\pi_{\mathrm{ref}}(y^- \mid x,k)}
% \right),
% \end{aligned}
% \]
% where $\pi_{ref}$ denotes the reference policy (we use the SFT-trained checkpoint in the experiments), $\beta$ scales preference impact. 
Training details are provided in Appendix~\ref{app:train}.

% \vspace{-5pt}
\section{Step-Level Human Preference Dataset}
\label{sec:dataset}
% \vspace{-5pt}
% \subsection{Dataset Schema}
% \label{sec:data_schema}

% Each data record corresponds to one cognitive trigger. Full decision context, response canditate sets, model self-ranking of the canditates, and human response are stored. Concretely, each record contains: 
% \begin{itemize}
%     \item \textbf{Decision context}: the prompt sent to the inference model, containing the observation, goal, retrieved memory, and current conversation history (if applicable).
%     \item \textbf{Candidate sets}: Candidate responses
%     \item \textbf{Model preference}: The LLM's preferred response 
%     \item \textbf{Human response}: Human's selected response or edited output
% \end{itemize}

% \begin{figure}[t]
%     \centering
%     \includegraphics[width=0.9\linewidth,height=0.27\textheight,keepaspectratio]{figures/dataset_sample.pdf}
%     \caption{An example of step-level human preference annotation in our dataset. 
% At each triggered cognitive step, the annotator is shown the decision context provided to the agent and multiple candidate outputs. 
% The annotator then selects the preferred output (green checkmark), producing one preference record grounded in the ongoing simulation.}
%     \label{fig:dataset_sample}
% \end{figure} 
\vspace{-6pt}
\subsection{Dataset Construction}
\vspace{-6pt}
Leveraging our \method \ interface and preference collection pipeline, we construct the first step-level human preference dataset for GA-style social simulation. 
The dataset comprises 57,239 preference pairs, spanning various social contexts with distinct personal goals under partial observability. 
Each data record corresponds to one triggered decision module $k\in\mathcal{K}$ for an agent at the simulation step, and is associated with the same LLM prompt at that moment (\Sref{sec:data_pipeline}). 

\begin{wraptable}{r}{0.55\columnwidth}
% \vspace{-1em}
\centering
\footnotesize
\setlength{\tabcolsep}{3pt}
\renewcommand{\arraystretch}{0.95}
\caption{\footnotesize Dataset composition and human--LLM alignment statistics. \#Pairs counts preference pairs per module. Pct. denotes the fraction of each module in the whole dataset. Top-1 Match denotes the fraction of triggers where the human choice matches the LLM top-ranked candidate. \#Cust. denotes the number of human-customized outputs in this module. Refl. Q. denotes the Reflection Question module. Imp. Sc. denotes the Importance Score module.}
\label{tab:ga_prefs}
\vspace{0.2em}
\begin{tabular}{lrrcr}
\toprule
Module & \#Pairs & Pct. & Top-1 Match & \#Cust. \\
\midrule
Plan & 2199 & 3.8 & 31.2 & 47 \\
Act & 19591 & 34.2 & 39.6 & 253 \\
Talk & 10573 & 18.4 & 35.3 & 593 \\
Reflect & 1142 & 2.0 & 47.1 & 0 \\
Refl. Q. & 1142 & 2.0 & 51.4 & 0 \\
Imp. Sc. & 22592 & 39.5 & 29.3 & 324 \\
\bottomrule
\end{tabular}
\vspace{-2.3em}
\end{wraptable}

Our dataset is constructed by eight software engineers with at least a bachelor's degree over a four-week period. 
Each decision step is annotated by a single annotator, with no overlap across annotators. 
To ensure annotation quality, annotators completed a calibration phase with supervised pilot simulations before formal annotation. 
The presentation order of the three candidates is randomized at each step to prevent annotators from defaulting to a fixed position. 
% Detailed data schema is provided in Appendix~\ref{app:data_schema}.

% \begin{table}[t]
% \centering
% \small
% \setlength{\tabcolsep}{4pt}
% \begin{tabular}{lrrrr}
% \toprule
% Component & \#Pairs & Percentage & Top-1 Match$\uparrow$ & \#Edits\\
% \midrule
% Plan & 2199 & 3.8 & $31.2$\% & 46\\
% Act & 19591 & 34.2 & $39.6$\% & 252\\
% Talk & 10573 & 18.4 & $35.3$\% & 593\\
% Reflect & 1142 & 2.0 & $47.1$\% & 0 \\
% Reflection Question & 1142 & 2.0 & $51.4$\% & 0\\
% Importance Score & 22592 & 39.5 & $29.3$\% & 325\\
% \bottomrule
% \end{tabular}
% \caption{Dataset composition and human--LLM alignment statistics. \#Pairs counts preference pairs per component. Percentage denotes the fraction of each component in the whole dataset. Top-1 Match denotes the fraction of triggers where the human choice matches the LLM top-ranked candidate. \#Edits denotes the number of human-edited outputs in this component.}
% \label{tab:ga_prefs}
% \end{table}

We summarize dataset composition in Table~\ref{tab:ga_prefs}. 
\texttt{Importance Score} is the most frequent module (39.5\%), followed by \texttt{Act} (34.2\%), \texttt{Talk} (18.4\%), \texttt{Plan} (3.8\%), \texttt{Reflection Question} (2.0\%), and \texttt{Reflect} (2.0\%). 
This distribution reflects the execution-heavy nature of long-horizon trajectories and the central role of memory in GA-style architectures.
\vspace{-2mm}

\vspace{-6pt}
\subsection{Human--LLM Preference Alignment}
\vspace{-6pt}
A unique feature of our dataset is that, for each triggered decision step, we record both the human-preferred output and the LLM's self-preference among the $3$ candidates presented to the annotator. 
We prompt the model to rank the generated 3 candidates according to its own preference during data collection. 
We quantify per-module preference agreement using \textbf{Top-1 Match}, defined as the fraction of outcomes in which the human choice matches the LLM's top-ranked candidate.
Candidate generation and self-preference ranking use the same step-level context exposed by the preference collection pipeline.
Since each trigger presents three candidates, the naive baseline for Top-1 Match under random agreement is about $33\%$.

We observe higher human--LLM agreement on the \texttt{Reflection Questions} and \texttt{Reflection} modules, with approximately 50\% of human choices matching the LLM's top choices. 
This higher agreement likely stems from the relatively formulaic nature of these modules and their reduced dependence on feasibility constraints. For \texttt{Act}, \texttt{Plan}, and \texttt{Talk}, human--LLM agreement drops substantially. 
These modules directly influence agent behavior and therefore require consistent grounding in the information state.

In particular, \texttt{Importance Score} has the smallest output space, which would inflate its chance agreement above the $\approx 33\%$ naive baseline; yet it exhibits the \emph{lowest} observed alignment of any module, with only $29.3\%$ of human choices matching the LLM's top-ranked candidate (Table~\ref{tab:ga_prefs}). 
This is notable because importance is not directly observable in agents' trajectory-level behaviors, yet it indirectly affects memory retrieval. 
Systematic miscalibration of importance scores could propagate downstream errors by promoting irrelevant memories and suppressing goal-critical ones, leading agents to forget arrangements or act inconsistently. 
We further connect this mismatch to behavior by ablating the importance estimator and examining this hypothesis in \Sref{sec:main_results}.
% \paragraph{Obedient and long dialogues hinder simulation quality.} While collecting the dataset, we observe that the base model GPT-4o rarely negates opinions and suggestions proposed by other agents in their conversations, despite the suggestions might not be appropriate. 
% We also observe that the base model is reluctant to finish conversation. 
% This is likely because GPT models are aligned to continue a topic as long as possible, and to be obedient to the user in conversations. 
% However, in agentic social simulations it hinders simulation quality due to long conversations with inefficient information diffusion.

\vspace{-10pt}
\section{Experiments}
\vspace{-10pt}
In this section, we aim to answer the following questions:
\begin{itemize}
    \item \textbf{Q1:} Can step-level preference learning improve open-weight LLMs on GA-style social event simulations?
    \item \textbf{Q2:} How do supervised finetuning and preference alignment, respectively, contribute to improvements in agent interaction quality and goal fulfillment? 
    \item \textbf{Q3:} How does step-level preference learning influence the agents' behavioral composition in long-horizon trajectories?
\end{itemize}

\vspace{-6pt}
\subsection{Experimental Setup}
\label{sec:exp_setup}
\vspace{-6pt}

\paragraph{Evaluation scenarios.} We evaluate \emph{scenario-level} generalization within the GA paradigm: preferences collected on the 30 training events are tested on 10 held-out social events with disjoint scripts, different agents, distinct goal assignments, and different information-asymmetry structures (\Sref{sec:dataset}).
Each event involves 4-8 agents with distinct goals, using the same simulator, perception pipeline, and action interface; methods differ only in the backbone model and/or the memory importance estimator. 
For each method and held-out event, we run 3 independent episodes each lasting 3 simulated days. 

\paragraph{Baselines and trained variants.}
We evaluate both proprietary and open-weight baselines. The proprietary baselines are GPT-4o \citep{openai_gpt4o_2024}, GPT-5.2 \citep{openai_gpt52}, and DeepSeek-v3.2 \citep{deepseek_v32_2025}. 
The open-weight baselines are Qwen2.5-7B-Instruct (Q7B), Qwen2.5-14B-Instruct (Q14B) \citep{yang2024qwen2.5}, and Llama-3.1-8B-Instruct (L8B) \citep{grattafiori2024llama3}. 
We further report a human reference consisting of trajectories generated by annotators controlling agents through \method\ under the same intervention setting as in data collection, judged by the same GPT-5.2 / DeepSeek-v3.2 evaluators as all other models.
To study the impact of step-level preference learning, we evaluate SFT models trained on accepted outputs. We then evaluate SFT+DPO models further trained on pairwise preference data, warm-starting from the SFT checkpoints. 
We also report \texttt{+MI} variants that keep the language model backbone unchanged and only replace the memory-importance estimator in retrieval scoring. The \texttt{+MI} variant is not intended as an apples-to-apples counterpart to full SFT/DPO; it isolates whether a drop-in improvement to the importance estimator alone, holding the backbone fixed, is sufficient to improve downstream behavior.
For retrieval-related embedding computations, we use \texttt{text-embedding-3-small} \citep{openai2024embedding3}.

\paragraph{Evaluation metrics.} We evaluate full trajectories using five event-centered metrics adapted from \citep{tian2025visualized}. For each trajectory, automated LLM evaluators are given the agents' goals, persona descriptions, and full execution histories and assign a 5-point Likert score for each metric. 
The five metrics are \textbf{Location Adherence} (whether the agent appears at the appropriate location for the intended activity), \textbf{Temporal Adherence} (whether actions occur at appropriate times relative to the event schedule), \textbf{Requirement Consistency} (whether behavior remains consistent with event constraints, available information, and agent profile), \textbf{Interaction Quality} (whether interactions are contextually relevant, coherent, and socially appropriate), and \textbf{Role Fulfillment} (whether behavior is consistent with the assigned persona and advances the explicit goal). 
We use GPT-5.2 \citep{openai_gpt52} and DeepSeek-v3.2 \citep{deepseek_v32_2025} as judges and report mean scores across evaluated trajectories in Table~\ref{tab:trajectory}.

\begin{table*}[t]
\centering
\scriptsize
\setlength{\tabcolsep}{2pt}
\renewcommand{\arraystretch}{1.05}
\caption{Trajectory performance scores across five evaluation dimensions. We compare base models, memory importance variants (MI), supervised fine-tuning (SFT), and direct preference optimization (DPO) methods, evaluated using GPT-5.2 and DeepSeek-v3.2 as judges (higher is better). \colorbox{gray!20}{Gray} indicates human reference. \textbf{Bold} denotes the best non-human performance. \textcolor{green!60!black}{$\uparrow$} / \textcolor{red!70!black}{$\downarrow$} denotes the absolute score increase / decrease from the base model. Overall, preference-aligned models (SFT and DPO) consistently outperform base models, with most of the gain coming from SFT; DPO adds a modest and uneven improvement over SFT, concentrated in location adherence and interaction quality, while approaching human-level performance in several metrics.}
\label{tab:trajectory}
\vspace{0.4em}
\begin{tabularx}{\textwidth}{@{}l *{5}{>{\centering\arraybackslash}X}@{}}
\toprule
\multirow{2}{*}{Model} &
Location &
Temporal &
Role &
Requirement &
Interaction \\
&
Adherence &
Adherence &
Fulfillment &
Consistency &
Quality \\
\midrule
\rowcolor{gray!20} Human & 3.12 & 3.07 & 3.25 & 3.41 & 3.09 \\
\midrule
GPT-4o        & 2.39 & 2.94 & 2.81 & 3.07 & 2.30 \\
GPT-5.2       & 2.68 & \textbf{2.97} & 3.05 & \textbf{3.09} & 3.01 \\
DeepSeek-v3.2 & 2.75 & 2.51 & 3.02 & 2.84 & 2.95 \\
Q7B-Base     & 1.51 & 1.71 & 2.13 & 1.79 & 1.95 \\
Q14B-Base     & 1.70 & 2.07 & 2.35 & 2.12 & 1.96 \\
L8B-Base     & 2.14 & 2.11 & 2.28 & 2.29 & 2.13 \\
\midrule
GPT-4o+MI &
2.38\textsuperscript{\tiny\textcolor{red!70!black}{$\downarrow$0.01}} &
2.77\textsuperscript{\tiny\textcolor{red!70!black}{$\downarrow$0.17}} &
2.79\textsuperscript{\tiny\textcolor{red!70!black}{$\downarrow$0.02}} &
3.06\textsuperscript{\tiny\textcolor{red!70!black}{$\downarrow$0.01}} &
2.39\textsuperscript{\tiny\textcolor{green!60!black}{$\uparrow$0.09}} \\
DeepSeek-v3.2+MI &
\textbf{2.88}\textsuperscript{\tiny\textcolor{green!60!black}{$\uparrow$0.13}} &
2.56\textsuperscript{\tiny\textcolor{green!60!black}{$\uparrow$0.05}} &
3.04\textsuperscript{\tiny\textcolor{green!60!black}{$\uparrow$0.02}} &
3.01\textsuperscript{\tiny\textcolor{green!60!black}{$\uparrow$0.17}} &
\textbf{3.17}\textsuperscript{\tiny\textcolor{green!60!black}{$\uparrow$0.22}} \\
Q7B+MI &
1.54\textsuperscript{\tiny\textcolor{green!60!black}{$\uparrow$0.03}} &
1.75\textsuperscript{\tiny\textcolor{green!60!black}{$\uparrow$0.04}} &
2.10\textsuperscript{\tiny\textcolor{red!70!black}{$\downarrow$0.03}} &
1.73\textsuperscript{\tiny\textcolor{red!70!black}{$\downarrow$0.06}} &
1.93\textsuperscript{\tiny\textcolor{red!70!black}{$\downarrow$0.02}} \\
Q14B+MI & 
1.77\textsuperscript{\tiny\textcolor{green!60!black}{$\uparrow$0.07}} &
2.13\textsuperscript{\tiny\textcolor{green!60!black}{$\uparrow$0.06}} &
2.42\textsuperscript{\tiny\textcolor{green!60!black}{$\uparrow$0.07}}&
2.24\textsuperscript{\tiny\textcolor{green!60!black}{$\uparrow$0.12}}& 
2.06\textsuperscript{\tiny\textcolor{green!60!black}{$\uparrow$0.10}} \\
L8B+MI & 
2.03\textsuperscript{\tiny\textcolor{red!70!black}{$\downarrow$0.11}} &
2.15\textsuperscript{\tiny\textcolor{green!60!black}{$\uparrow$0.04}} &
2.26\textsuperscript{\tiny\textcolor{red!70!black}{$\downarrow$0.02}}&
2.27\textsuperscript{\tiny\textcolor{red!70!black}{$\downarrow$0.02}}& 
2.08\textsuperscript{\tiny\textcolor{red!70!black}{$\downarrow$0.05}} \\
\midrule
Q7B-SFT &
2.27\textsuperscript{\tiny\textcolor{green!60!black}{$\uparrow$0.76}} &
2.76\textsuperscript{\tiny\textcolor{green!60!black}{$\uparrow$1.05}} &
2.81\textsuperscript{\tiny\textcolor{green!60!black}{$\uparrow$0.68}} &
2.93\textsuperscript{\tiny\textcolor{green!60!black}{$\uparrow$1.14}} &
2.60\textsuperscript{\tiny\textcolor{green!60!black}{$\uparrow$0.65}} \\
Q14B-SFT &
2.21\textsuperscript{\tiny\textcolor{green!60!black}{$\uparrow$0.51}} &
2.48\textsuperscript{\tiny\textcolor{green!60!black}{$\uparrow$0.41}} &
2.85\textsuperscript{\tiny\textcolor{green!60!black}{$\uparrow$0.50}} &
2.55\textsuperscript{\tiny\textcolor{green!60!black}{$\uparrow$0.43}} &
2.35\textsuperscript{\tiny\textcolor{green!60!black}{$\uparrow$0.39}} \\
L8B-SFT &
2.37\textsuperscript{\tiny\textcolor{green!60!black}{$\uparrow$0.23}} &
2.75\textsuperscript{\tiny\textcolor{green!60!black}{$\uparrow$0.64}} &
2.84\textsuperscript{\tiny\textcolor{green!60!black}{$\uparrow$0.56}} &
2.93\textsuperscript{\tiny\textcolor{green!60!black}{$\uparrow$0.64}} &
2.64\textsuperscript{\tiny\textcolor{green!60!black}{$\uparrow$0.51}} \\
Q7B-SFT+DPO &
2.33\textsuperscript{\tiny\textcolor{green!60!black}{$\uparrow$0.82}} &
2.70\textsuperscript{\tiny\textcolor{green!60!black}{$\uparrow$0.99}} &
2.83\textsuperscript{\tiny\textcolor{green!60!black}{$\uparrow$0.70}} &
2.89\textsuperscript{\tiny\textcolor{green!60!black}{$\uparrow$1.10}} &
2.64\textsuperscript{\tiny\textcolor{green!60!black}{$\uparrow$0.69}} \\
Q14B-SFT+DPO &
2.41\textsuperscript{\tiny\textcolor{green!60!black}{$\uparrow$0.71}} &
2.62\textsuperscript{\tiny\textcolor{green!60!black}{$\uparrow$0.55}} &
2.74\textsuperscript{\tiny\textcolor{green!60!black}{$\uparrow$0.39}} &
2.63\textsuperscript{\tiny\textcolor{green!60!black}{$\uparrow$0.51}} &
2.47\textsuperscript{\tiny\textcolor{green!60!black}{$\uparrow$0.51}} \\
L8B-SFT+DPO & 
2.75\textsuperscript{\tiny\textcolor{green!60!black}{$\uparrow$0.61}} &
2.56\textsuperscript{\tiny\textcolor{green!60!black}{$\uparrow$0.45}} &
\textbf{3.08}\textsuperscript{\tiny\textcolor{green!60!black}{$\uparrow$0.80}} &
2.91\textsuperscript{\tiny\textcolor{green!60!black}{$\uparrow$0.62}} &
2.71\textsuperscript{\tiny\textcolor{green!60!black}{$\uparrow$0.58}} \\
\bottomrule
\end{tabularx}
\end{table*}

\vspace{-6pt}
\subsection{Main Results}
\label{sec:main_results}
\vspace{-6pt}

\paragraph{Step-level preference learning narrows the performance gap.} Table~\ref{tab:trajectory} shows that proprietary models (GPT-4o, DeepSeek-v3.2 and GPT-5.2) outperform open-weight baselines across all five metrics, indicating a substantial capability gap in social simulation. 
However, step-level preference learning markedly improves open-weight models. 
Compared with their base versions, both SFT and SFT+DPO consistently improve trajectory quality across all metrics, with especially large gains in temporal adherence, requirement consistency, and role fulfillment. 

\paragraph{Supervised finetuning accounts for most of the gains.}
Compared with the base models, most gains already appear after supervised finetuning on human-preferred module outputs.
This indicates that many simulation failures stem from poor local decisions at individual steps, rather than from a lack of general linguistic ability. 
Training on accepted step-level outputs helps the model produce actions and dialogues that are more aligned with agent roles and social constraints. 
As a result, agents become better at coordinating with timing, adhering to scenario requirements, and behaving consistently with their backgrounds and goals.
\vspace{-2pt}

\paragraph{Preference optimization brings modest and uneven additional gains.}
Because SFT already trains on the accepted outputs $y^+$, the added value of DPO is the explicit contrast against the rejected $y^-$, and its effect is correspondingly smaller and dimension-dependent.
Averaged over the three open-weight models, DPO improves location adherence ($+0.21$) and interaction quality ($+0.08$) over SFT, is essentially flat on role fulfillment ($+0.05$) and requirement consistency ($+0.01$), and is slightly lower on temporal adherence ($-0.04$).
The largest gains appear where several candidates are individually plausible but differ in grounding to the current information state or social appropriateness: by explicitly contrasting preferred and rejected outputs, preference optimization helps the model distinguish better choices from merely plausible ones.

\paragraph{Improving memory importance alone is insufficient.}
The \texttt{+MI} results show that improving only the memory importance estimator yields noticeably smaller and less consistent gains than full step-level alignment. 
This finding suggests that retrieval is only one part of the agent decision pipeline. 
Even with higher-ranked memories, the agent may still fail to plan appropriately, interact coherently, or choose actions that advance the event. 
Stronger long-horizon social behavior appears to require coordinated improvement across multiple modules, rather than isolated optimization of memory retrieval alone.

\begin{figure}[t]
    \centering
    \includegraphics[trim=5mm 5mm 5mm 5mm, clip,width=0.95\linewidth]{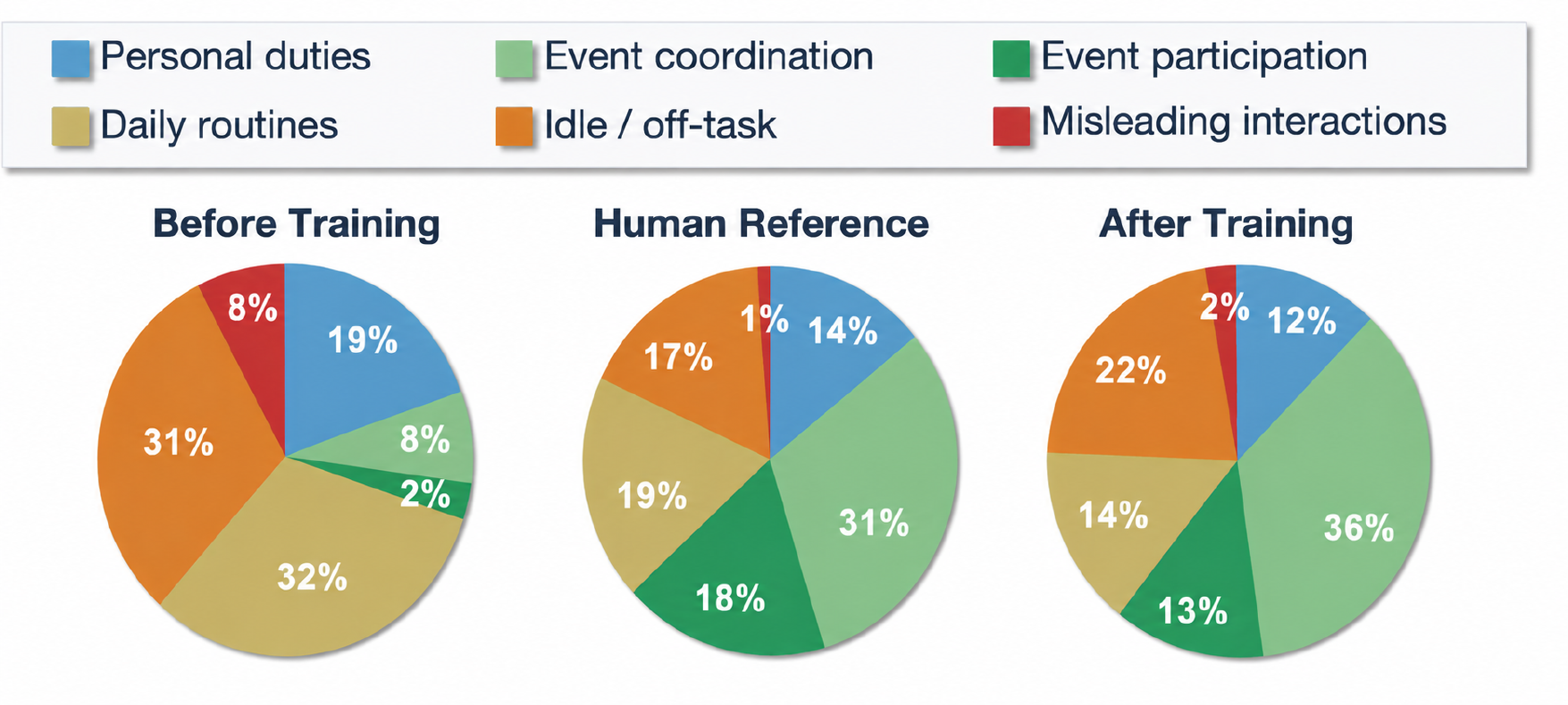}
    \vspace{-2mm}
    \caption{
    Distribution of agent time allocation across behavioral categories 
    before training, after training, and in human reference data. 
    After step-level preference learning, agents spend more time on event coordination 
    (+28pp) and event participation (+11pp), while reducing daily routines (-18pp), 
    idle/off-task behavior (-9pp), and misleading interactions (-6pp). 
    The resulting distribution more closely matches human behavior.
    }
    \label{fig:behavior_distribution}
\end{figure}
\vspace{-3mm}

\subsection{Behavioral Distribution Analysis}
\label{sec:behavior_distribution}
\vspace{-6pt}

To answer Q3, we analyze how step-level preference learning changes the long-horizon behavioral composition of agents. 
We analyze the distribution of time spent across behavioral categories of Q14B-Base and Q14B-SFT+DPO, and compare these distributions with human reference data. 

We categorize agent activities into six types: 
(i) personal duties, 
(ii) daily routines, 
(iii) event coordination, 
(iv) event participation, 
(v) idle/off-task behavior, 
and (vi) misleading interactions. 
Figure~\ref{fig:behavior_distribution} shows the proportion of time allocated to each category.

\paragraph{Improved role alignment and reduced harmful behaviors.}
After training, agents exhibit a clear shift toward role-aligned behaviors. 
Time spent on event coordination and participation become more balanced. 
In particular, event coordination increases substantially from 8\% to 36\%, and event participation also increases substantially from 2\% to 13\%.  
We observe a decrease in undesirable behaviors: 
idle/off-task behavior is reduced from 31\% to 22\%, 
and misleading interactions drop from 8\% to 2\%. 
These results indicate that step-level preference learning encourages agents to engage more actively in socially meaningful activities and provides effective signals to discourage irrelevant or harmful actions.

\paragraph{Closer alignment with human reference distribution.}
Overall, the post-trained distribution more closely resembles the human reference. 
We further quantify distributional alignment by computing the KL divergence over the normalized frequencies of behavior categories. 
The divergence to the human reference decreases from 0.610 before training to 0.084 after training, confirming that step-level preference learning improves not only individual decisions, but also the overall long-horizon composition of agent behavior.
Taken together, these long-horizon results address the concern that step-level (local) supervision might induce myopic or collapsing agents: over 3-day simulations the behavioral distribution moves toward the human reference and coordination and participation rise, rather than degrading, indicating that local preference signals translate into coherent global behavior.

\begin{figure}[t!]
    \centering
    \includegraphics[width=\linewidth]{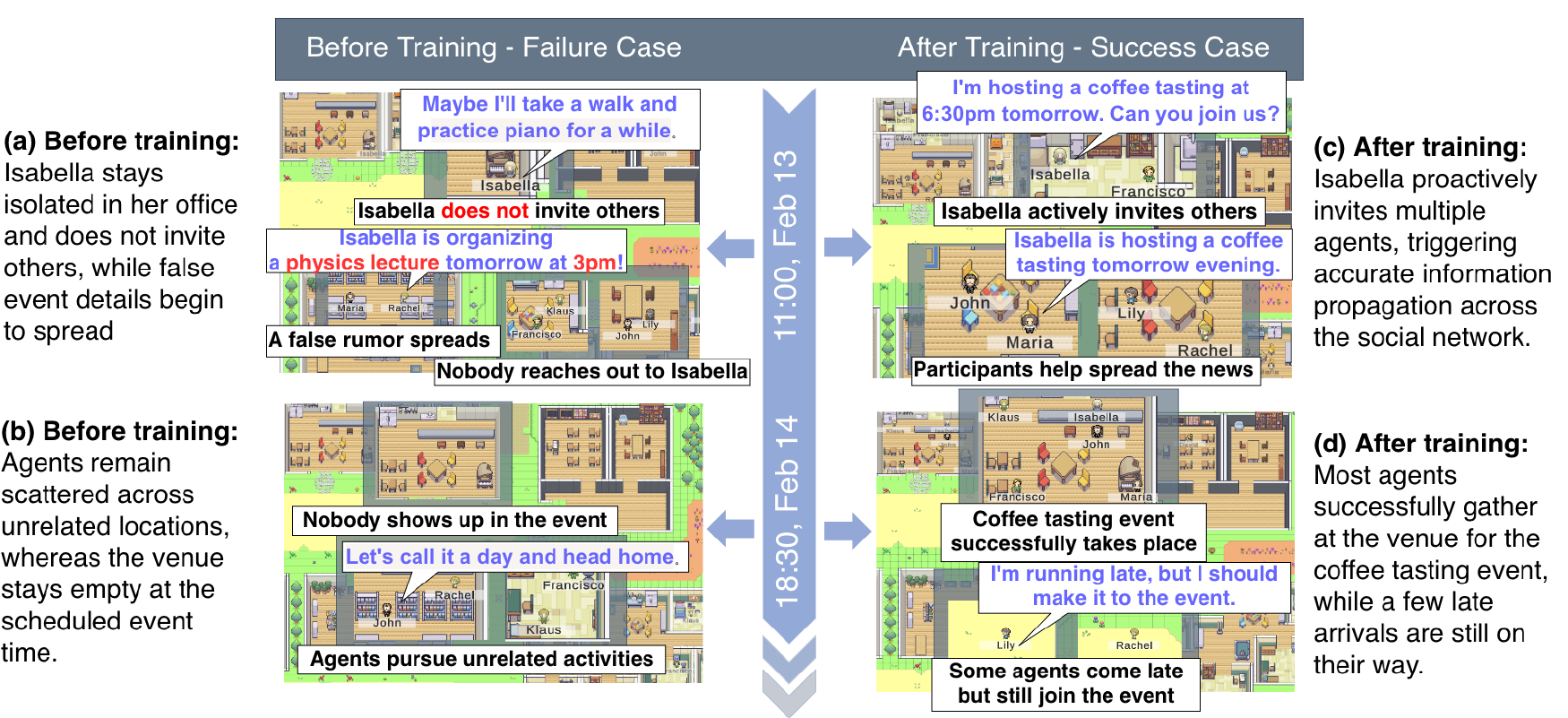}
    \caption{Case study of event information propagation before and after preference training. Panels (a)-(b) show that before training, agents seldom interact and spread a false rumor, and the event does not take place. Panels (c)-(d) show that after training, agents actively exchange correct event information, forming a successful invitation chain and enabling the event to take place.}
    \vspace{-10pt}
    \label{fig:case_study}
\end{figure}

\vspace{-6pt}
\subsection{Case Studies}
\label{sec:case_study}
\vspace{-6pt}

To complement our quantitative analysis, we examine how step-level preference finetuning affects agent behavior in concrete simulation episodes.
This case study is a qualitative complement, not independent statistical evidence: its role is to illustrate the mechanism behind the aggregate results, and the population-level evidence it illustrates is the behavioral shift in Figure~\ref{fig:behavior_distribution} and the KL-to-human drop from 0.610 to 0.084 (\Sref{sec:behavior_distribution}).
Figure~\ref{fig:case_study} contrasts the same social event before and after training at two stages: information propagation (Feb. 13) and event execution (Feb. 14). 
In this event, Isabella is responsible for hosting a coffee tasting event on Feb. 14 at 3:00 PM in her cafe. 
Other agents do not know the event details and are expected to gather the missing information and then participate in the event.

\paragraph{Case study 1: Broken information propagation before preference training.} Before training, the key coordinator, Isabella, remains isolated and does not initiate invitations, while another agent generates and spreads incorrect event details (Figure~\ref{fig:case_study}(a)). The resulting information breakdown leads to complete event failure, with no agents arriving at the venue (Figure~\ref{fig:case_study}(b)). 

\paragraph{Case study 2: Successful event coordination after preference training.} After training, Isabella proactively disseminates event information, whereas other agents help propagate it after meeting Isabella (Figure~\ref{fig:case_study}(c)). As a result, most participants successfully attend the event (Figure~\ref{fig:case_study}(d)). 
These patterns are consistent with the quantitative gains in location adherence, temporal adherence, and role fulfillment scores reported in Table~\ref{tab:trajectory}. 

% Overall, this case study suggests that preference training improves not only local response quality, but also the global dynamics of information propagation and multi-agent coordination.

% Overall, these case studies provide qualitative evidence that preference training improves local information exchange and leads to more effective multi-agent coordination in social event simulation. 
% These qualitative observations are consistent with the quantitative improvements reported in Table~\ref{tab:trajectory}, particularly in Temporal Adherence and Role Fulfillment, where trained models show the largest gains.
\enlargethispage{2\baselineskip}

\vspace{-10pt}
\section{Limitations and Future Work}
\label{sec:limitations}
\vspace{-10pt}

Our work has several limitations. First, although we collect step-level human preferences across multiple modules, current data is collected from a finite set of event scenarios with a limited number of annotators, which may not fully capture broader variations in social norms, interaction styles, or long-horizon coordination strategies. Second, each decision step is labeled by a single annotator without overlap, so we do not report inter-annotator agreement; while our annotation targets decision competence rather than subjective social taste and uses a supervised calibration phase and randomized candidate order (\Sref{sec:dataset}), the absence of a measured agreement statistic remains a limitation. Third, our evaluation relies primarily on LLM-as-a-judge scoring, which, although far more affordable than human annotation, carries some degree of evaluator bias relative to human scoring. Fourth, the collected data is heavily module-imbalanced, with two modules accounting for over $70\%$ of all pairs, which can bias the jointly trained preference signal toward high-frequency modules.

Future work can extend this line of research in several directions. A natural next step is to expand the dataset to more diverse scenarios, more heterogeneous annotator populations, and more varied social goals. It is also worth characterizing the marginal value of the annotation effort through a data-quantity ablation, for example training on subsets of the accepted and preference data to test whether performance saturates before the full 57K pairs, and verifying that step-level alignment preserves general capabilities by evaluating on general-purpose benchmarks such as MMLU or IFEval before and after training. Beyond the scenario-level generalization we study, cross-environment and cross-architecture transfer remains open. Finally, combining step-level human preferences with trajectory-level supervision is a promising direction for jointly optimizing internal module decisions and global social outcomes.

\vspace{-10pt}
\section{Conclusion}
\label{sec:conclusion}
\vspace{-10pt}

In this work, we introduce \method, an interactive social simulation interface that enables efficient step-level human supervision of the outcomes of agent internal modules. 
Leveraging the interface, we construct the first human preference dataset conditioned on GA-style simulations. 
We perform step-level preference learning on open-weight base models and demonstrate consistent improvements on held-out events across event-coordination and role-fulfillment metrics.

% Beyond performance gains, MA-Prefs enables fine-grained diagnosis of where current LLM agents diverge from human preferences.
% We find that human--LLM agreement is substantially higher for reflection-related components than for planning, action, and dialogue, and that memory-importance estimation exhibits particularly strong miscalibration despite its small discrete label space.
% Our ablations further suggest that aligning \emph{indirect} components (especially importance) can improve rollouts by reshaping retrieved context, while aligning \emph{direct} components improves immediate execution and grounded interaction compliance.
Overall, our results highlight the value of step-level human preferences as both training signals and analytical tools for social simulation agents. 
We hope \method \ will facilitate more principled evaluation and alignment of fine-grained agent decisions as a complement to trajectory-level evaluation, with the combination of step-level and trajectory-level supervision a promising direction for future work.

\vspace{-10pt}

% \clearpage

\section*{Acknowledgements}
\vspace{-10pt}
All content was originally written by the authors; 
generative AI tools were used solely for language polishing 
and grammar checking.

\raggedbottom
\bibliographystyle{splncs04}
\bibliography{colm2026_conference}

\vspace{10pt}

\appendix
\enlargethispage{8\baselineskip}
\par\noindent{\large\bfseries Appendix}\par
\vspace{-0.8em}
\nopagebreak[4]
\renewcommand{\theHsection}{appendix.\Alph{section}}
\renewcommand{\theHsubsection}{appendix.\Alph{section}.\arabic{subsection}}
% \vspace{-10pt}
\section{Interface Illustration} 
\label{app:interface}
\vspace{-25pt}
\FloatBarrier

\begin{figure}[H]
\centering
\captionsetup{skip=2pt}
\includegraphics[width=0.58\textwidth]{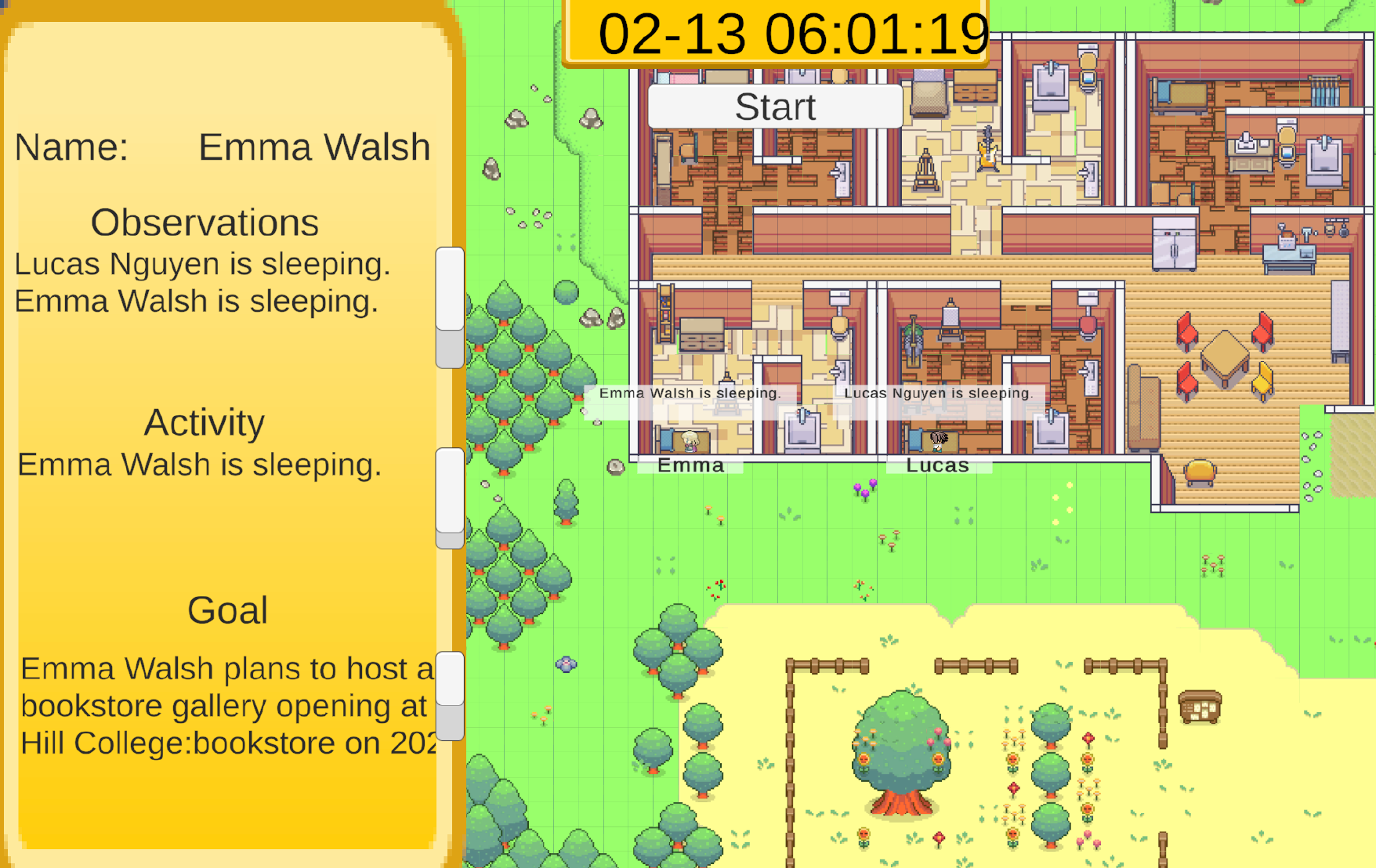}
\caption{Overview of the \method \ interface. The left panel shows controlled-agent observations, activity, and goal, while the map view shows agents and their current activities; annotators can switch agents or inspect other agents during simulation.}
\label{fig:interface_overview}
\end{figure}

\vspace{50pt}

\begin{figure}[H]
    \centering
    \captionsetup{skip=2pt}

    \begin{subfigure}[t]{0.32\textwidth}
        \centering
        \includegraphics[width=\linewidth,height=0.10\textheight]{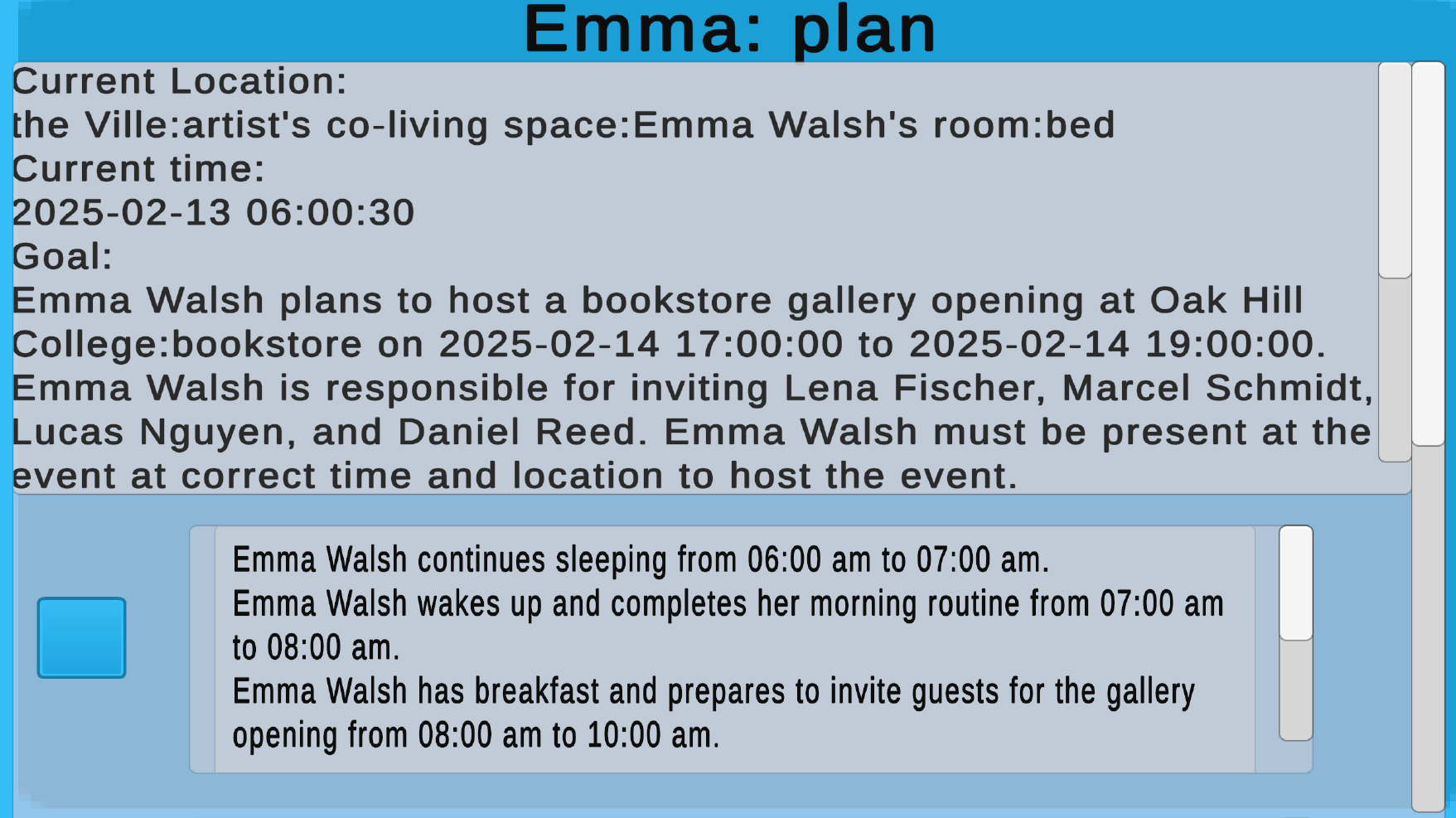}
        \caption{Decision context}
        \label{fig:appendix_interface_overview}
    \end{subfigure}
    \hfill
    \begin{subfigure}[t]{0.32\textwidth}
        \centering
        \includegraphics[width=\linewidth,height=0.10\textheight]{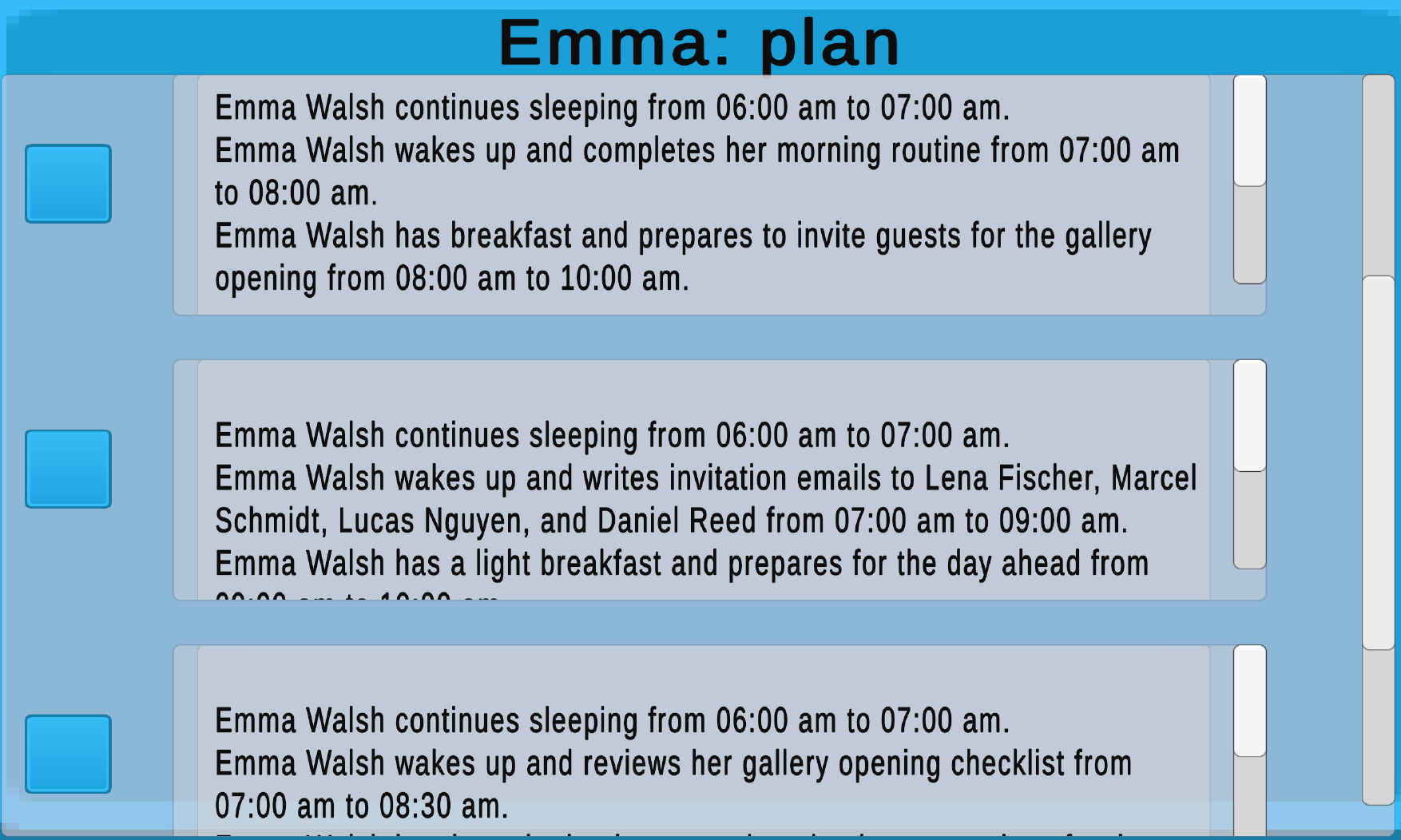}
        \caption{Candidate selection}
        \label{fig:appendix_interface_context}
    \end{subfigure}
    \hfill
    \begin{subfigure}[t]{0.32\textwidth}
        \centering
        \includegraphics[width=\linewidth,height=0.10\textheight]{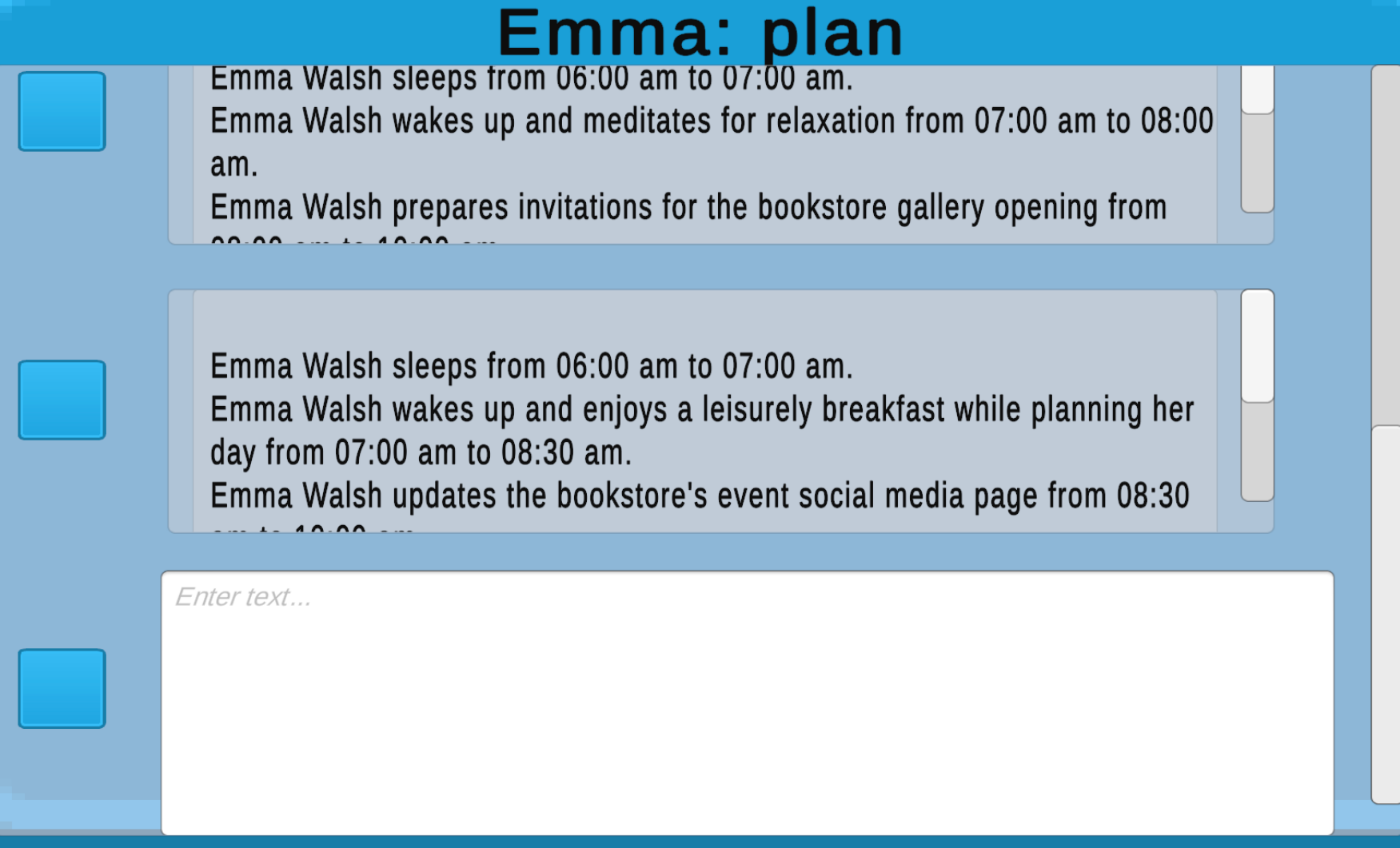}
        \caption{Output customization}
        \label{fig:appendix_interface_candidates}
    \end{subfigure}

    \caption{Screenshots of the \method \ annotation interface}
    \label{fig:app_annotation_interface}
\end{figure}

\vspace{-20pt}
\section{Additional Training Details}
\label{app:train}
\vspace{-10pt}

% WAICA only
LoRA (rank 16, $\alpha=16$, dropout 0.05) is applied to the self-attention projection matrices (\texttt{q}, \texttt{k}, \texttt{v}, \texttt{o}) in both SFT and SFT+DPO stages of Qwen2.5-7B-Instruct and Llama-3.1-8B-Instruct experiments. For Qwen2.5-14B-Instruct experiments, we applied LoRA with the same hyperparameters to self-attention projection matrices {\texttt{q}, \texttt{v}}. 
We use AdamW with weight decay 0.1, warmup ratio 0.05, gradient clipping 1.0, bf16 mixed precision, max sequence length 4096, and train for 1 epoch on 2$\times$A100 GPUs. 
SFT uses a cosine schedule with learning rate $3\times10^{-5}$ for Llama and $2\times10^{-5}$ for Qwen. 
DPO ($\beta=0.05$) uses a linear schedule with learning rate one order of magnitude smaller ($3\times10^{-6}$ and $2\times10^{-6}$, respectively), warm-started from the SFT checkpoint.

\vspace{-10pt}
\section{Implementation Details}
\label{app:details}
\vspace{-5pt}

% WAICA2026 only
\subsection{Interface and Engine}
\method \ follows a server--client architecture: a backend engine maintains a centralized world state and a discrete simulation clock over a tile-based spatial representation built on the SmallVille assets~\citep{park2023generative}, while clients visualize the simulation for human annotators. We instantiate 60 agent profiles across 40 social events (30 for training, 10 for evaluation). To accommodate the additional latency of generating multiple candidates for protagonist agents ($\approx$5s vs.\ $\approx$1s for normal agents), the engine applies time dilation only to protagonist agents during data collection.

Each agent runs in its own worker thread and communicates with the engine through a thread-safe queue, enabling concurrent cognition while keeping world updates deterministic. Agents output structured actions in a restricted space---navigation, descriptive activities, and dialogue initiation---each accompanied by an observable event description and an estimated duration. New actions are generated only when the current activity elapses, and can be preempted by incoming events or dialogue.

\vspace{-6pt}
\subsection{Event Design}
\label{sec:event_design}
\vspace{-6pt}

% WAICA2026 only
Each event centers on a shared social objective involving a \textit{host} with near-complete event information and \textit{participants} who observe only partial details, each assigned an explicit goal $g_i$ describing their responsibility. 
This asymmetry forces agents to acquire missing details through interaction or memory retrieval---naturally inducing confirmation, invitation, coordination, and occasional miscommunication---and admits multiple valid trajectories, yielding the diverse candidate outputs needed for step-level annotation.

\end{document}